\documentclass[a4paper,conference]{IEEEtran}
\usepackage[T1]{fontenc}
\usepackage{color}
\usepackage{float}
\usepackage{mathrsfs}
\usepackage{amsmath}
\usepackage{amssymb}
\usepackage{graphicx}
\usepackage{float} 

\usepackage{booktabs} 
\usepackage{caption}  
\usepackage{CJK}
\usepackage{indentfirst}
\usepackage{amsmath}
\usepackage{cases}
\usepackage{setspace}

\usepackage[unicode=true,
bookmarks=true,bookmarksnumbered=true,
bookmarksopen=true,bookmarksopenlevel=1,
breaklinks=false,pdfborder={0 0 0},
pdfborderstyle={},backref=false,
colorlinks=false]
{hyperref}
\hypersetup{pdftitle={Your Title},
pdfauthor={Han Yulu},
pdfpagelayout=OneColumn, 
pdfnewwindow=true, 
pdfstartview=XYZ, 
plainpages=false}
\makeatletter

\floatstyle{ruled}
\newfloat{algorithm}{tbp}{loa}
\providecommand{\algorithmname}{Algorithm}
\floatname{algorithm}{\protect\algorithmname}

\usepackage[caption=false,font=footnotesize]{subfig}
\usepackage{algorithm}
\usepackage{algorithmic}

\usepackage{multirow} 
\usepackage{amsmath} 
\usepackage{xcolor}

\allowdisplaybreaks[4]

\ifCLASSOPTIONcompsoc
\usepackage[caption=false,font=normalsize,labelfont=sf,
textfont=sf]{subfig}
\else
\usepackage[caption=false,font=footnotesize]{subfig}
\fi

\usepackage{cite}
\usepackage{bm}
\usepackage{algorithmic}
\usepackage{algorithm}
\usepackage{graphicx}
\IEEEoverridecommandlockouts

\usepackage{lettrine}

\usepackage{geometry}
\usepackage{subfig}
\makeatother
\usepackage{geometry}
\usepackage{subfig}
\usepackage{hyperref}
\makeatother
\begin{document}

\title{\textcolor{black}{CNN+Transformer Based 
Anomaly Traffic Detection in UAV Networks for Emergency Rescue}}
\author{
\IEEEauthorblockN{Yulu Han$^{\dagger}$, Ziye Jia$^{\dagger}$, Sijie He$^{\dagger}$, Yu Zhang$^{\S}$, Qihui Wu$^{\dagger}$
\\
 }
\IEEEauthorblockA{
$^{\dagger}$The Key Laboratory of Dynamic Cognitive System of Electromagnetic Spectrum Space, Ministry of Industry and\\ 
Information Technology, Nanjing University of Aeronautics and  Astronautics, Nanjing, Jiangsu, 211106, China\\
$^{\S}$Department of Electronic and Electrical Engineering, University College London, London WC1E 6BT, United Kingdom\\
\{hanyulu, jiaziye, sx2304109, wuqihui\}@nuaa.edu.cn, uceezhd@ucl.ac.uk
}

\thanks{{This work was supported in part by the Natural Science Foundation on Frontier Leading 
Technology Basic Research Project of Jiangsu under Grant BK20222001,  
in part by National Natural Science Foundation of China under Grant 62301251, 
in part by the Aeronautical Science Foundation of China 2023Z071052007, and in part by the Young 
Elite Scientists Sponsorship Program by CAST 2023QNRC001.
 }}

}
\maketitle

\thispagestyle{empty}
\begin{abstract}
 The unmanned aerial vehicle (UAV) network has 
 gained significant attentions in recent years due to its various applications. 
 However, the traffic security becomes the key 
 threatening public safety issue in an emergency rescue 
 system due to the increasing vulnerability of UAVs to cyber attacks in 
 environments with high heterogeneities. 
 Hence, in this paper, we propose a novel anomaly traffic detection architecture for 
 UAV networks based on the software-defined networking (SDN) framework and blockchain technology. 
 Specifically, 
 SDN separates the control and data plane to enhance the network manageability and security.
 Meanwhile, the blockchain provides decentralized identity authentication 
 and data security records. 
 Beisdes, a complete security architecture requires an 
 effective mechanism to detect the time-series based abnormal traffic. 
 Thus, an integrated algorithm combining 
 convolutional neural networks (CNNs) and Transformer (CNN+Transformer) for anomaly traffic detection is 
 developed, which is called CTranATD.  
 Finally, the simulation results show that the proposed CTranATD algorithm is effective 
 and outperforms the individual CNN, Transformer, and LSTM algorithms for 
 detecting anomaly traffic.

\end{abstract}

\newcommand{\CLASSINPUTtoptextmargin}{0.8in}

\newcommand{\CLASSINPUTbottomtextmargin}{1in}

\section{Introduction}
\vspace{2mm}
\lettrine[lines=2]{D}UE to the mobility, convenience, driverless, 
and high flexibility, unmanned aerial vehicles (UAVs) 
play important roles in both military and 
civilian fields, such as 
disaster management, search and rescue,  
data collection, and monitoring\cite{1,2,3}. 
When UAVs implement an environmental monitoring mission to 
improve the emergency rescue system,  
frequent UAV data interactions, dynamic network topologies, 
and high heterogeneity of devices bring risks of potential cyber attacks, such as Hijacking, Denial-of-service
 (DoS), and Distributed Denial of Service (DDoS) \cite{4,5,6}. 
These attacks can lead to abnormal traffic, causing the failure of the UAV 
missions and threatening public safety. Therefore, 
it is siginificant to design an anomaly traffic detection model 
for UAVs to protect the communication safety. 

There exist a couple of works utilizing machine learning and deep learning techniques 
for anomaly detection. For instance, authors in \cite{7} 
leverage the support vector machine 
and isolation based 
active learning techniques as unsuperivised anomaly detection methods, which 
separate the groups of data using density based spatial 
clustering of applications. 
\cite{8} developes a time-series fault diagnosis model based on the 
deep transformer technology, which improves the accuracy 
of lithium battery fault diagnosis. 
In \cite{9}, the authors propose a customizable and communication-efficient 
federated anomaly detection scheme by combining the temporal convolutionl 
network attention mechanisms and a 
federated learning framework to facilitate abnormal log patterns. 
For the security in UAVs, authors in \cite{10} propose deep learning techniques 
of CNN, long short-term memory (LSTM), and the combination of CNN and LSTM (CNN+LSTM) models.
To classify network intrusions and attacks, \cite{11} utilizes adaptive synthetic 
sampling with three maching learning models of the K-nearest neighbor, 
decision tree, and random forest. 

However, as far as the authors' knowledge, the existed works ignore the 
communication security architecture for UAV networks and the 
anomaly traffic detection based on time series transmitted by 
UAVs. 
The software-defined networking (SDN) is a revolutionary architecture that separates the 
control and data planes, offering advantages such as
enhanced manageability, flexibility and network programmability,
as well as the ability to introduce new solutions to
address security threats \cite{12,13}. 
Besides, the blockchain technology has been applied to many fields due to the benefits of 
decentralization, data transparency, and traceability, making it 
a powerful tool to solve the problems of efficiency and 
security in traditional systems \cite{14}. 
Additionally, the advantages of 
the CNN extracting local features and the 
ability of the Transformer analyzing the time-series data in the long term 
trends make the CNN+Transformer mechanism suitable to detect abnormal traffic. 
Therefore, we design a CNN+Transformer algorithm for anomaly traffic detection called CTranATD, 
which is deployed on a safe communication 
architecture with the combination of the SDN and blockchain to protect the 
communication safety with UAVs in emergency rescues.

In this work, we focus on enhancing the performance of the anomaly traffic detection 
and construct a security model to provide identity authentication and data security records
for UAV networks. A security architecture based on the SDN and blockchain 
is designed to manage data, which ensures the data security when  
UAVs collecing and forwarding data. Further, 
an anomaly traffic detection algorithm based on the CTranATD 
is proposed, which is deployed on the SDN control plane to 
identify and detect abnormal data and time-series traffic collected and 
forwarded by UAVs. To verify the performance, we conduct extensive trainings 
and predictions 
based on the CICIDS2017 dataset \cite{15}. Simulation results demonstrate the effective performance of 
the proposed CTranATD in the anomaly traffic detection. 

The rest of this paper is arranged as follows. 
Section \ref{section2} presents the system model.
In Section \ref{section3}, an anomaly traffic detection model 
based on CTranATD is designed.
we present the results and analyses in Section \ref{section4}. 
Finally, Section \ref{section5} draws conclusions.
\vspace{0.2cm}

\section{System Model}\label{section2}
\vspace{2mm}
\subsection{Scenario Description}

\begin{figure}[t]
     \centering
     \includegraphics[width=0.68\linewidth]{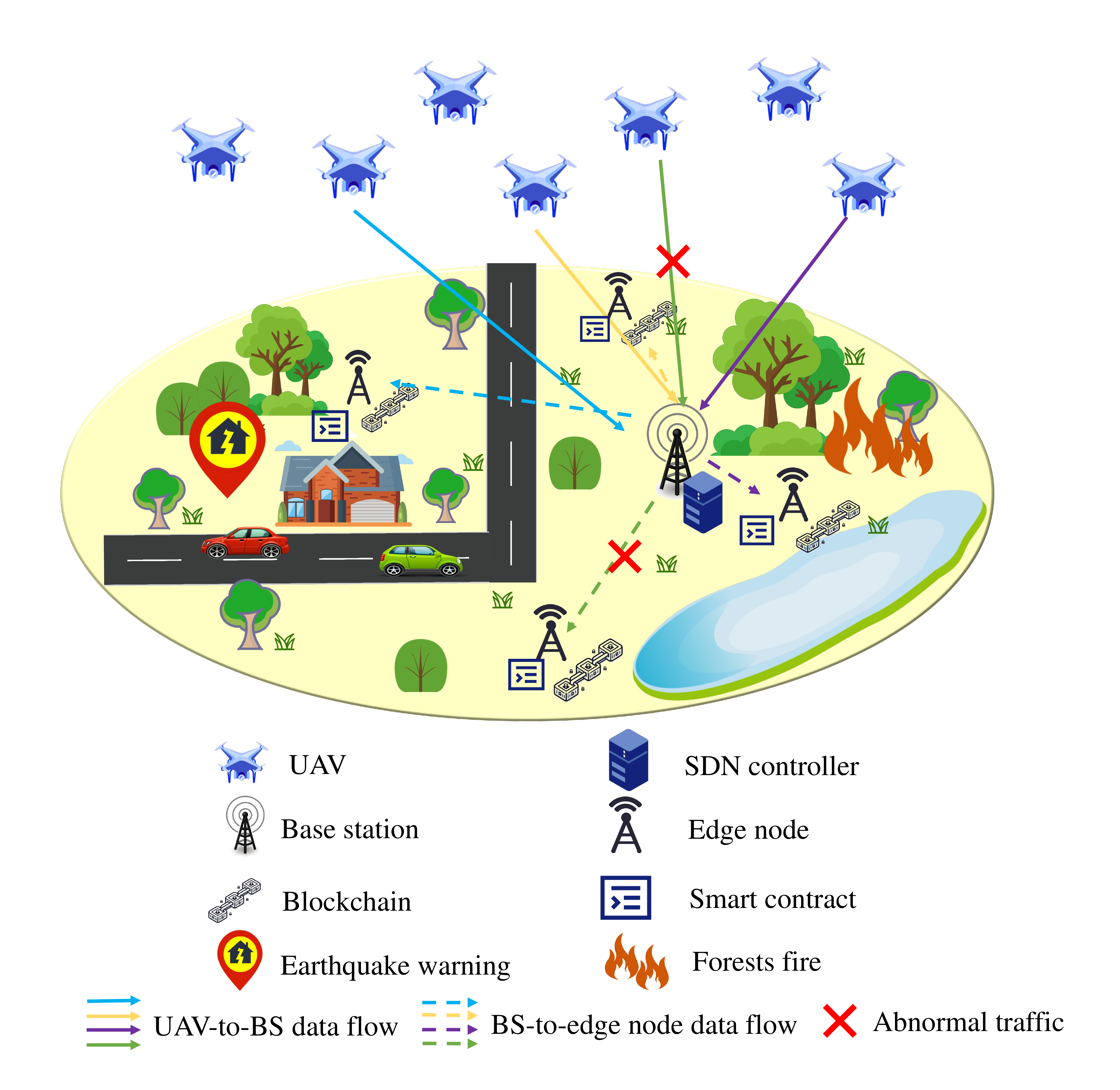}
     \captionsetup{font=small} 
     \caption{\raggedright A communication security network for
     UAVs combining the SDN and blockchain in emergency rescue}
     \label{fig:Scene}
\end{figure}

As shown in Fig. \ref{fig:Scene},  there exist several UAVs equipped with 
various sensors in the air to monitor and collect 
the natural disasters related information for emergency rescue, 
which is significantly related with public safety. 
The information includes the surface acceleration, surface gravity, 
atmosphere pressure, humidity, and other environmental data from both cities and suburbs.
UAVs need to transmit the data to the central base station (BS) 
timely, so that the data can be forwarded to the distributed edge nodes 
for processing. The BS does not participate in the process of data and 
is equipped with a central switch with a SDN controller. 
An anomaly traffic detection algorithm is deployed on the SDN controller 
to monitor and classify traffic. 
After detecting the traffic, the SDN controller relays the data to the edge nodes. 
In the proposed model of Fig. \ref{fig:Scene}, 
the solid lines represent data transmission from the UAV 
to BS, while the dashed lines represent transmissions 
from the SDN control plane to the edge nodes. 
Each UAV transmitts different types of data to certain edge 
nodes for processing. The abnormal traffic is marked with crosses when the UAV 
is attacked. This paper mainly focuses on detecting three attacks including DoS, DDoS, and PortScan in UAV communicatin networks, 
which can lead to network congestion, data errors, and even information loss. These threats are  
inconducive to the emergency rescue, networks, and social security.

\subsection{SDN and Blockchain Based Security Model}
To protect the UAV network from cyber attacks, we design a 
SDN and blockchain based security model. The SDN controller transmitts the data 
after detection by the CTranATD. 
The blockchain is distributed on every edge node to 
record and update identities, flight logs, and the data states of UAVs. 
The cooperation between the SDN and blockchain is supported by smart contracts, 
which are self-executing protocols running on a blockchain utilizing programmable languages 
to realize automation. 
In the proposed model, the SDN controller works collaboratively with the 
blockchain to build a secure, efficient, and traceable UAVs communication 
environment through efficient data management and transparent recording mechanisms. 

The SDN controller is responsible for the traffic monitoring 
of the global network with its centralized architecture to receive data transmitted by 
UAVs. Additionally, the SDN controller deployed with the anomaly 
traffic detection algorithm analyzes the traffic quickly and labels 
them as ``norma'' or ``abnormal''. Smart contracts take over and perform 
subsequent actions after receiving the results sent by the SDN controller. 
As for the normal traffic, the smart contracts update the identity 
information and flight logs of the UAV on the chain to ensure the 
legitimacy, while recording the environmental disaters concerning data for traceability. 
As for the abnormal traffic, the smart contracts update the identity information 
and abnormal flight logs on the chain, and notify the SDN controller 
for the prohibition of transmission. Meanwhile, as a distributed 
ledger, the blockchain utilizes miners to package and store packets 
passing the detection. The traceability and immutability of the data are 
realized by the blockchain through these records, providing 
a guarantee for the subsequent identification of responsibility. 

However, an effecitive anomaly traffic detection algorithm is the key to 
detecting abnormal traffic and maintaining the SDN and blockchain based 
security model. Thus, the CTranATD deployed on the SDN controller is proposed 
to settle the binary classification problem.

\vspace{2mm}
\section{Algorithm Design}\label{section3}
\begin{figure}
     \centering
     \includegraphics[width=0.66\linewidth]{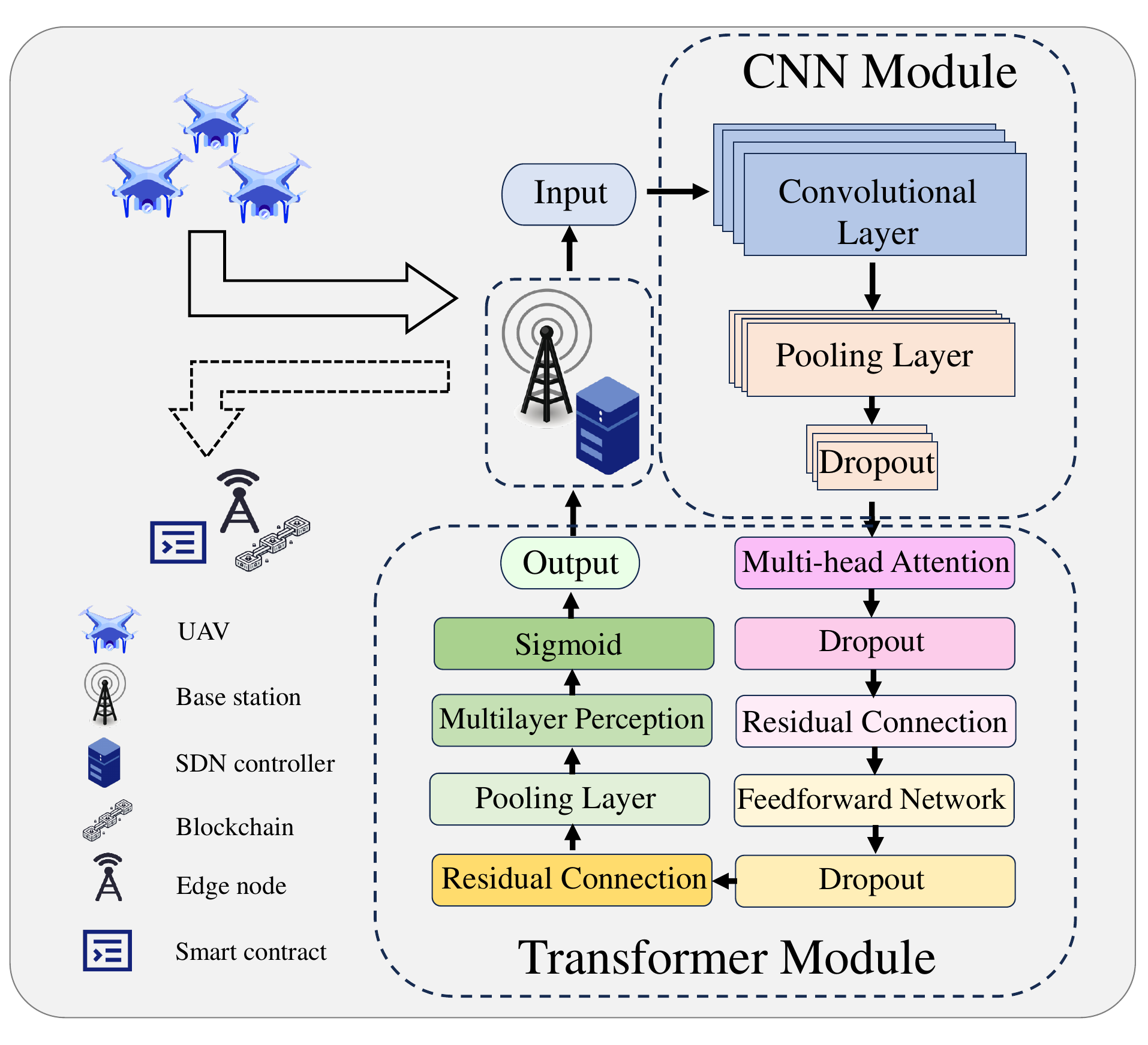}
     \captionsetup{font=small}
     \vspace{-2mm}
     \caption{\raggedright The CTranATD proposed for anomaly traffic detection based on time series.}
     \label{fig:algorithm}
\end{figure}

\vspace{2mm}
The CTranATD algorithm combining the CNN and Transformer 
modules is deployed on the SDN controller to conduct 
the anomaly traffic detection, as is shown in 
Fig. \ref{fig:algorithm}. The data transmitted by UAVs is put 
into the CNN module before entering Transformer. 
The convolutional layer, pooling layer, and  
dropout layer are designed sequentially in the CNN. 
In the Transformer module, multi-head attention mechanisms, residual connections, feedforward 
networks, pooling layers, and multilayer perceptions are set to train and 
learn the characteristics of the data. 
After the detection by CTranATD, the SDN controller sends the 
results to smart contracts on the blockchain deployed on the edge nodes.
CICIDS2017 is leveraged as the training dataset including  
different kinds of features representing the traffic state.
However, data of three attacks need to be preprocessed 
before training. 

\subsection{Data Preprocessing}
Considering the superior ability of Transformer to process 
time-series data and the need to reduce redundant contents, we process the 
data according to the time stamp. 
We assume that UAVs transmit environmental data collected  
by sensors and traffic states once per second. The 60 pieces of data are selected 
randomly per minute to form each dataset for prediction.

The contents of data preprocessing also contain standardizing 53 features from the 
original dataset, including total forward packets, total backward packets, etc. 
Besides, source Internet protocol (IP) and destination 
IP are converted into numeric form by using hash encoding. 
The source and destination ports are normalized with the min-max normalization function. 
The label "protocol" is mapped to an integer with protocol class tag 
for the convenience of processing by the CTranATD.

Considering the mission for emergency rescue during natural disasters, 
we add the atmosphere pressure, humidity, and other environmental 
information consistent with the traffic state of the data for analyzing 
natural ecological activities and changes. The data corresponding to 
the abnormal state caused by cyber attacks  
may exceed the fluctuation range of normal data, which have 
adverse effects on emergency rescues. 
\vspace{-2mm}

\subsection{CTranATD Algorithm for Anomaly Traffic Detection}

Firstly, the preprocessed data is put into a CNN network
which handles one dimension sequences with the shape of 3D tensor. 
The content is the batch size, window size, and feature dimensions. 
It represents the feature data in a 
time window, which is set as 60, to 
comply with the processing of time series based CTranATD and the 
data transmitted by UAVs. 
There are 71 feature dimensions after processing.

The CNN can reduce the length of features through convolution 
and pooling operations, which is 
\begin{equation}
     y_{\mathrm{conv}}(t,f)=\mathrm{ReLU}\left(\sum_{i=1}^kw_{f,i}\cdot x(t+i-1)+b_f\right),
\end{equation}
and
\begin{equation}
     y_{\mathrm{pool}}(t,f)=\max\left(y_{\mathrm{conv}}(t+i,f)\right),\quad\forall i\in\mathrm{pool_{size}},
\end{equation}
where $x(t)$ represents the input seuqence, $w_{f,i}$ denotes 
the weight of $f$ filter, $k$ represents the size of kernel, 
$b_{f}$ is the bias of the convolution kernel, $t$ is the time step, 
$f$ denotes the index of filters, and $y_\mathrm{conv}(t,f)$ and $y_\mathrm{pool}(t,f)$ 
are the outputs of convolutional layer and pooling layer, respectively. 
The two layers help retain important 
local information and extract short term trends and the changes 
of patterns. $\mathrm{ReLU}\left(\cdot\right)$ denotes the activation function used in 
the CNN module.

In detail, the convolution layer extracts local features 
with sliding windows, and each convolution kernel 
learns specific set of features. The detection algorithms of 
three attacks in this paper all uses 64 convolution cores. 
The size of kernel is fine tuned acccording to each 
data of attacks. 
Thus, the output of CNN module is 
\begin{equation}
 \begin{aligned}
    &y_{\mathrm{output}}(t,f)=\mathrm{Dropout}\\
    &\!\left(\!\max_{i\in\mathrm{pool}_\mathrm{size}}\!\left(\mathrm{ReLU}\left(\sum_{j=1}^kw_{f,j}\cdot x(t\!+\!j\!-\!1)+b_f\!\right)\!\right)\!\right),
 \end{aligned}
\end{equation}
where $\mathrm{Dropout}\left(\cdot\right)$ denotes the action to drop neurons.

However, although CNN is good at extracting local features and 
reducing computational complexity, it is insufficient for 
analyzing time series when it comes to the long-term trends. 
Therefore, the output from the dropout layer of the CNN module 
enters into the multi-head attention layer of Transformer 
for feature learning, which forms the CTranATD algorithm.
The multi-head self-attention mechanisms in the encoder of Transformer 
can be used to capture 
the dependencies from different location in time series\cite{16}. 
The function of attention weights is 
\begin{equation}
     Attention(Q,K,V)=softmax\left(\frac{QK^{T}}{\sqrt{d_{k}}}\right)V,
\end{equation}
in which query vector $Q$ is calculated from the word embedding and 
weight matrix of the input sequence. Key vector $K$ is calculated 
from the input sequence and weight matrix. Value vector $V$ is 
generated by the input sequence. The $softmax$ function 
normalizes the correlations to probability distributions. 
$\sqrt{d_k}$ represents the scaling factor preventing excessive dot product values 
from gradient disappearance.

The multi-head attention concatenates the output of each head. 
Then, the final output is intergrated by a linear transformation, i.e., 
\begin{equation}
     \mathrm{MultiHead}(Q,\!K,\!V)\!=\!\mathrm{Concat}(\mathrm{head}_1,\ldots,\mathrm{head}_h)\!W_O,
\end{equation}
in which $h$ denotes the number of heads, $W_O$ represents the output 
transformation matrix after multi-head attentions, and $\mathrm{Concat}\left(\cdot\right)$ means concatenation.

The residual connection can retain the original information and 
superimpose multi-head attention mechanisms to avoid the loss of 
information at the same time. The feedforward network is laid between 
each step as a fully connected layer, which can further transform 
features. Then, the nonlinear representation ability  
is improved by the ReLU activation function. 
The multilayer perception can further extract and map after learning. 
The sigmoid activation function is used to output the prabability of 
positive class.
We utilize the loss fuction to measure the difference 
between the predicted result and real label, and the classification task 
is finally completed.

The steps of preprocessing is the same when predicting. 
The predicting results include the 
accuracy, recall and other index parameters to measure the quality of 
classification.

\begin{table}[htbp]
    \centering
    \caption{Parameters Set} 
    \vspace{-2mm}
        \begin{tabular}{ l c c c} 
            \toprule
            Attacks & DoS & DDoS & PortScan \\ 
            \midrule
            CNN kernel & 5 & 3 & 3 \\ 
            Pool size & 2 & 2 & 2 \\
            Head size & 4 & 4 & 4 \\ 
            Head number & 2 & 2 & 2 \\ 
            Feadforward dimension & 64 & 64 & 64 \\
            Transformer block & 1 & 1 & 1 \\
            Multilayer perception & 64 & 64 & 64 \\
            Dropout & 0.1 & 0.1 & 0.1 \\
            Batch size & 32 & 32 &32 \\
            \bottomrule
        \end{tabular}
    \label{tab:model parameters}
\end{table}
\begin{figure}
     \centering
     \includegraphics[width=0.64\linewidth]{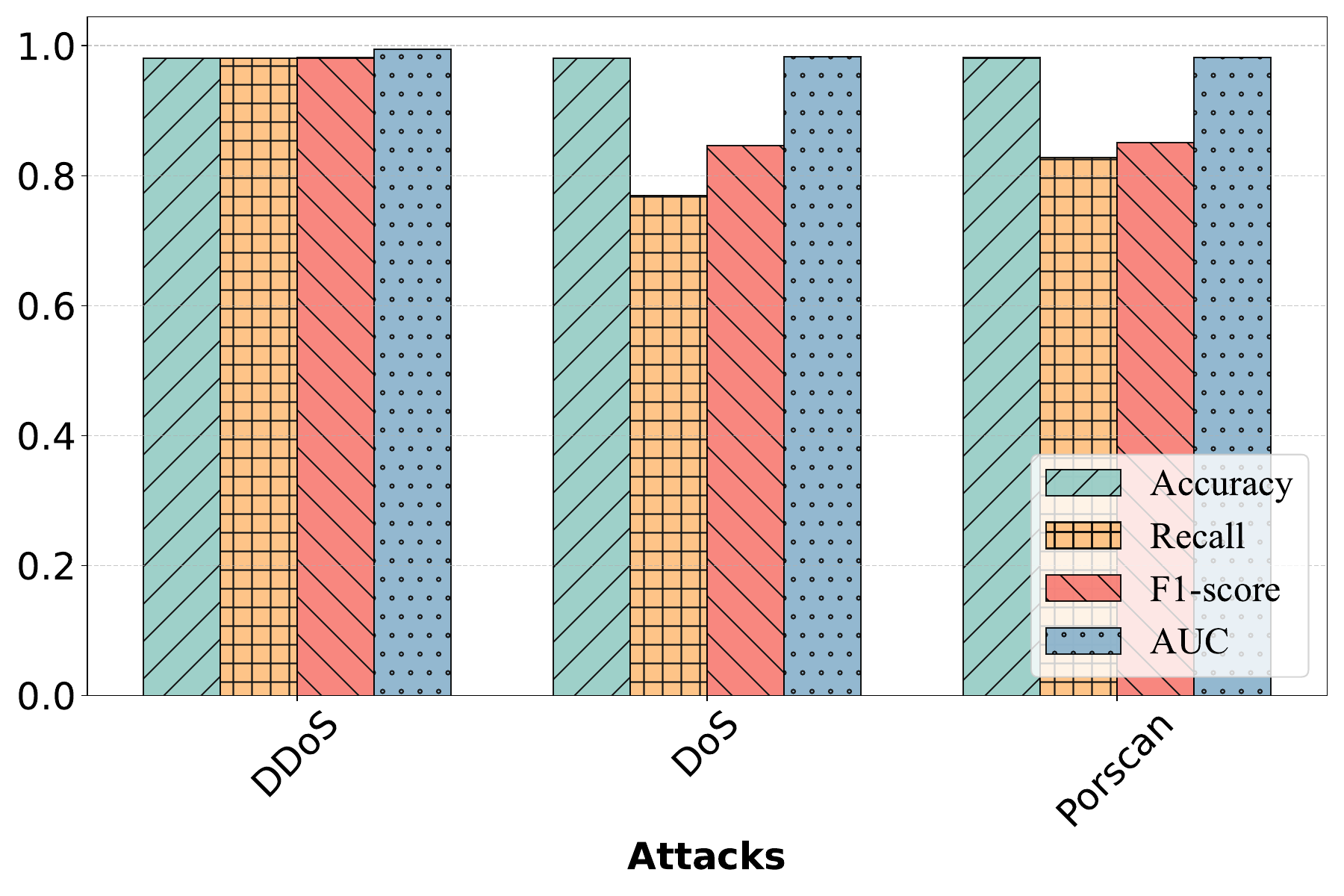}
     \captionsetup{font=small}
     \caption{\raggedright Performance of 3 Attacks.}
     \label{fig:results}
\end{figure}
\section{Simulation and Analysis}\label{section4}
\vspace{2mm}
We leverage the GPU of RTX4090D (24 GB) with the environment of 
Pytorch 2.3.0, Python 3.12, and Cuda 12.1 for training. 
All the indexes are dimensionless measurements between 0 and 1. 
In detail, there exist four situations. The positive class represents the abnormal state of traffic, and the
negative class represents the opposite.
True Positive (TP) and False Negative (FN) represent that the positive class is predicted to be positive and negative, respectively.
Meanwhile, 
the False Positive (FP) and True Negative (TN) represent that the negative class is predicted
to be positive and negative, respectively.
In this paper, several evaluation indexes of binary classification 
problems are used to measure the prediction ability of anomaly detection 
algorithms, including Accuracy, Recall, F1-score, Receiver Operating Characteristic (ROC) curve, 
and the Area Under the Curve of ROC (AUC), detailed as follows.

\begin{itemize}
     \item Accuracy is the most intuitive index to evaluate the 
classification correctness of the model, representing the 
percentage of the total number of samples that the algorithm predicted correctly. 
It can be expressed as
\begin{equation}
     \mathrm{Accuracy}=\frac{\mathrm{TP}+\mathrm{TN}}{\mathrm{TP}+\mathrm{TN}+\mathrm{FP}+\mathrm{FN}}.
\end{equation}

\item Recall measures the ability of the algorithm to identify positive samples, i.e.,
\begin{equation}
     \mathrm{Recall}=\frac{\mathrm{TP}}{\mathrm{TP}+\mathrm{FN}}.
\end{equation}

\item The F1-score is the reconciled average of precision and recall, and the formula is
\begin{equation}{\label{}}
    \mathrm{F1-Score}=2\cdot\frac{\mathrm{Precision}\cdot\mathrm{Recall}}{\mathrm{Precision}+\mathrm{Recall}},
\end{equation}
where $\mathrm{Precision}$ represents the proportion of all samples predicted 
to be positive that are truely positive. 

\item ROC curves indicate the classification ability under different thresholds. 
The horizontal coordinate of the ROC curves is the false positive 
rate (FPR), and the vertical coordinate is the true positive rate (TPR). 

\item AUC is a numeric measure of the ability to distinguish 
positive and negative samples. A higher AUC indicates the better  
ability of the algorithm to predict and classify the samples. 

\end{itemize}

Table \ref{tab:model parameters} shows the specific training parameters 
of each algorithm under different attacks. Parameters are the same 
in Transformer layer except for some minor adjustments in CNN kernels. 
Such parameter settings ensure better classification results of algorithms while 
lower the training time and reduce the computing rescources. 
\begin{figure}
	\centering
	\subfloat[Different performances of PortScan.]
      {
	\includegraphics[width=0.58\linewidth]{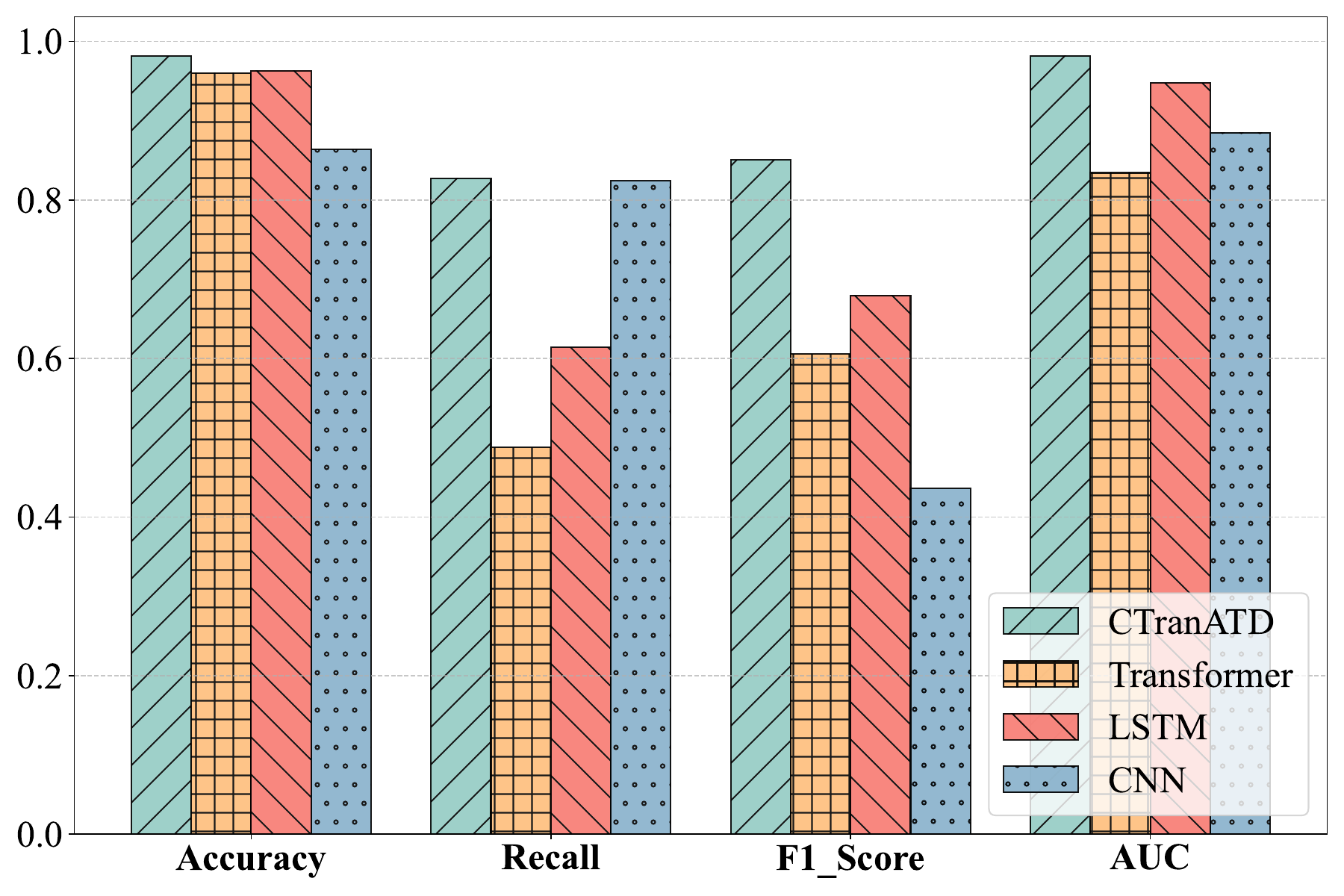}
     \label{fig:3a}
      } 
     \vspace{-2mm}
    \quad 
    \subfloat[Different performances of DoS.]
     {
	\includegraphics[width=0.58\linewidth]{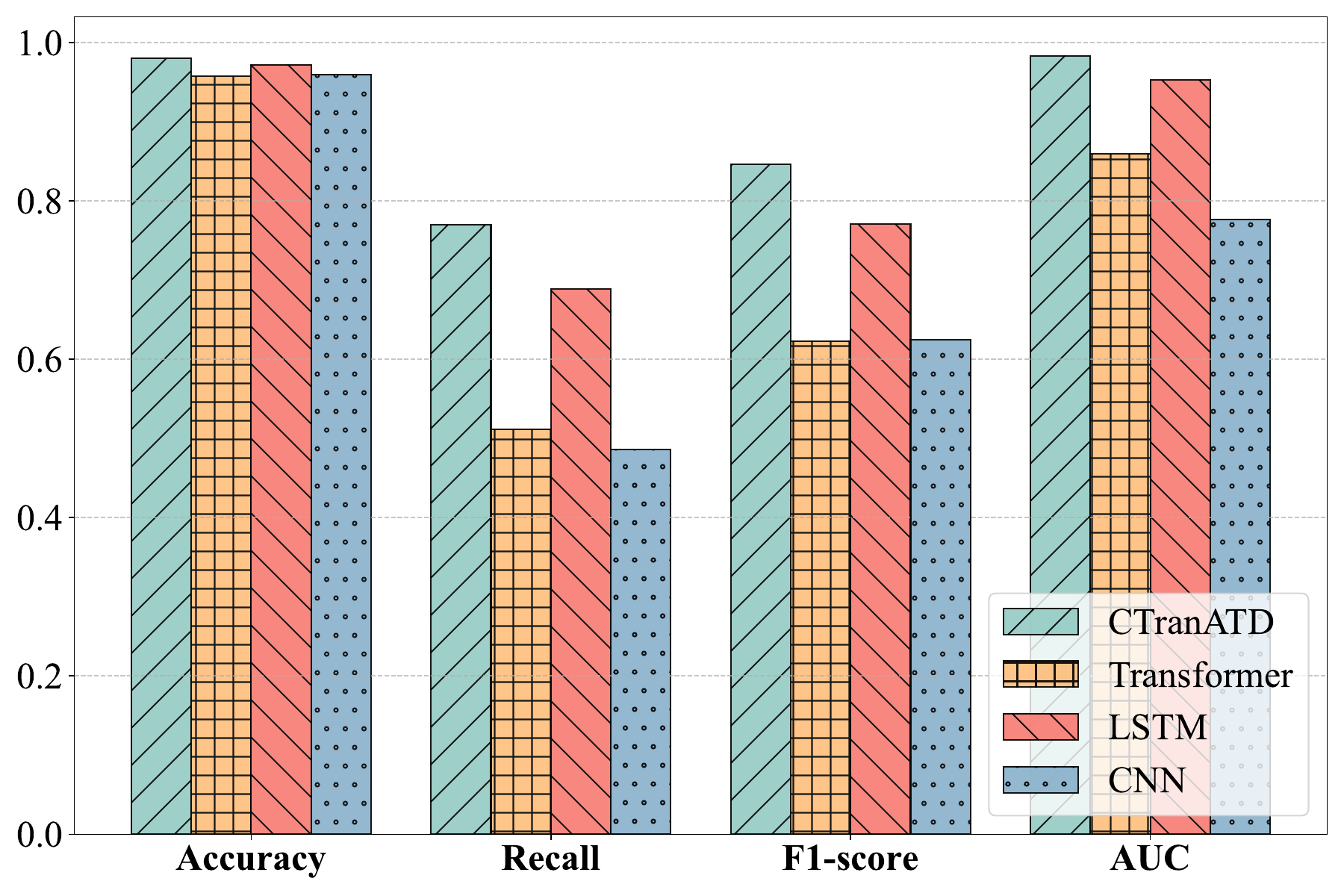}
     \label{fig:3b}
     }
     \vspace{-2mm}
     \quad
     \subfloat[Different performances of DDoS.]
     {
	\includegraphics[width=0.58\linewidth]{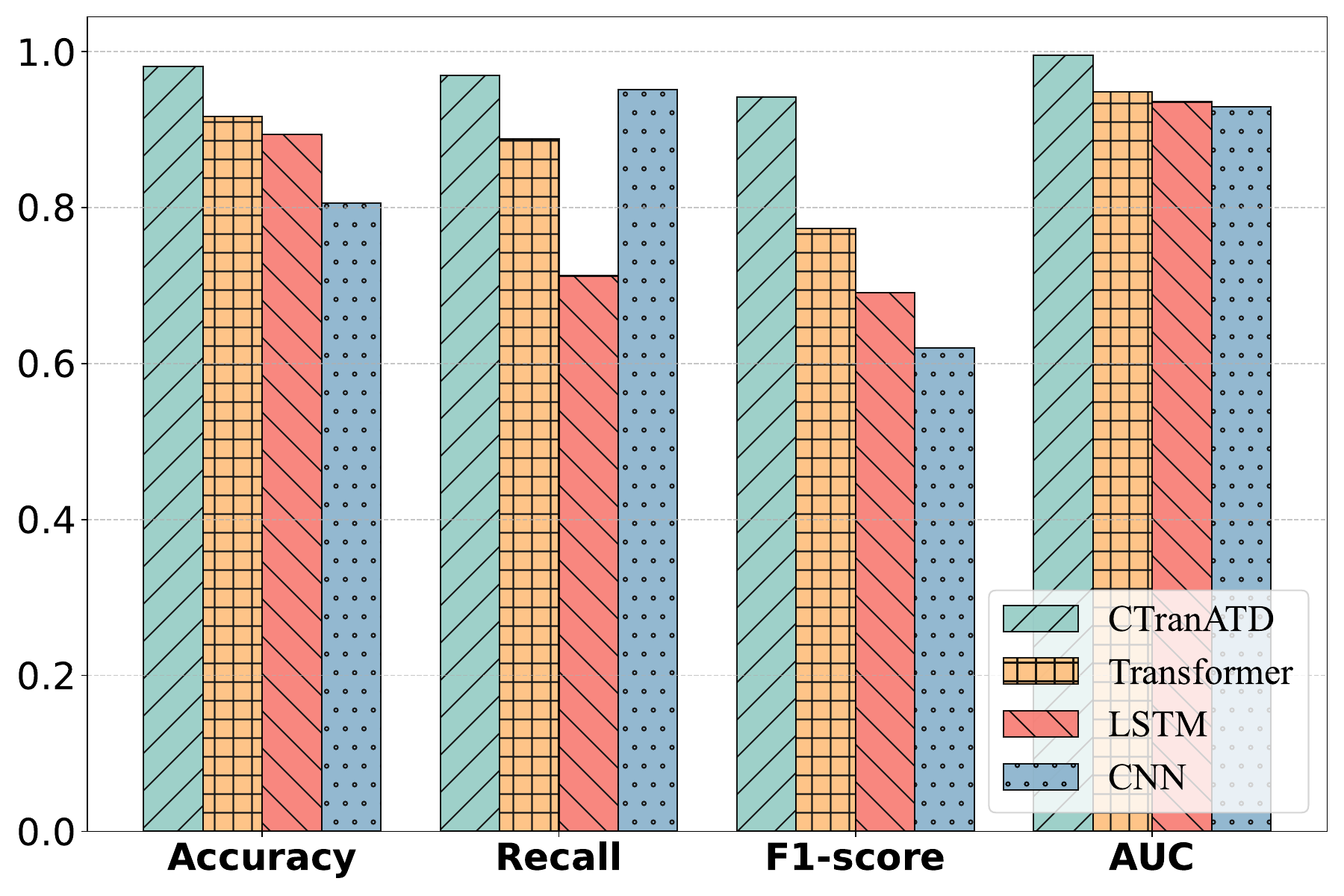}
     \label{fig:3c}
     }
     \captionsetup{font=small}
	\caption{\raggedright Performance of different algorithms with different attacks.}
	\label{fig:dif met}
\end{figure}
\begin{figure}
	\centering
	\subfloat[ROC curves for PortScan.]
      {
	\includegraphics[width=0.60\linewidth]{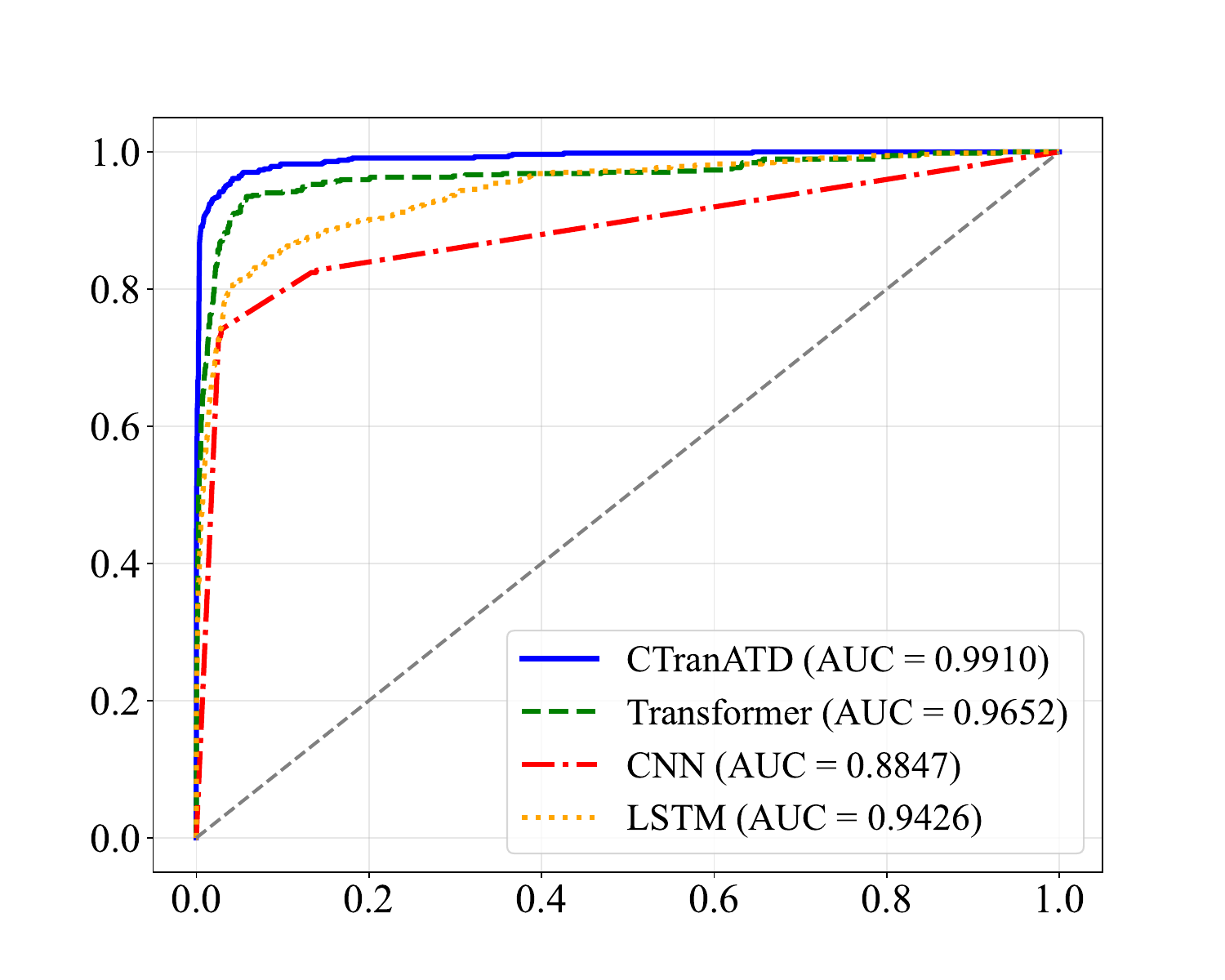}
     \label{fig:4a}
      } 
     \vspace{-4mm}
     \quad 
     \subfloat[ROC curves for DoS.]
     {
	\includegraphics[width=0.60\linewidth]{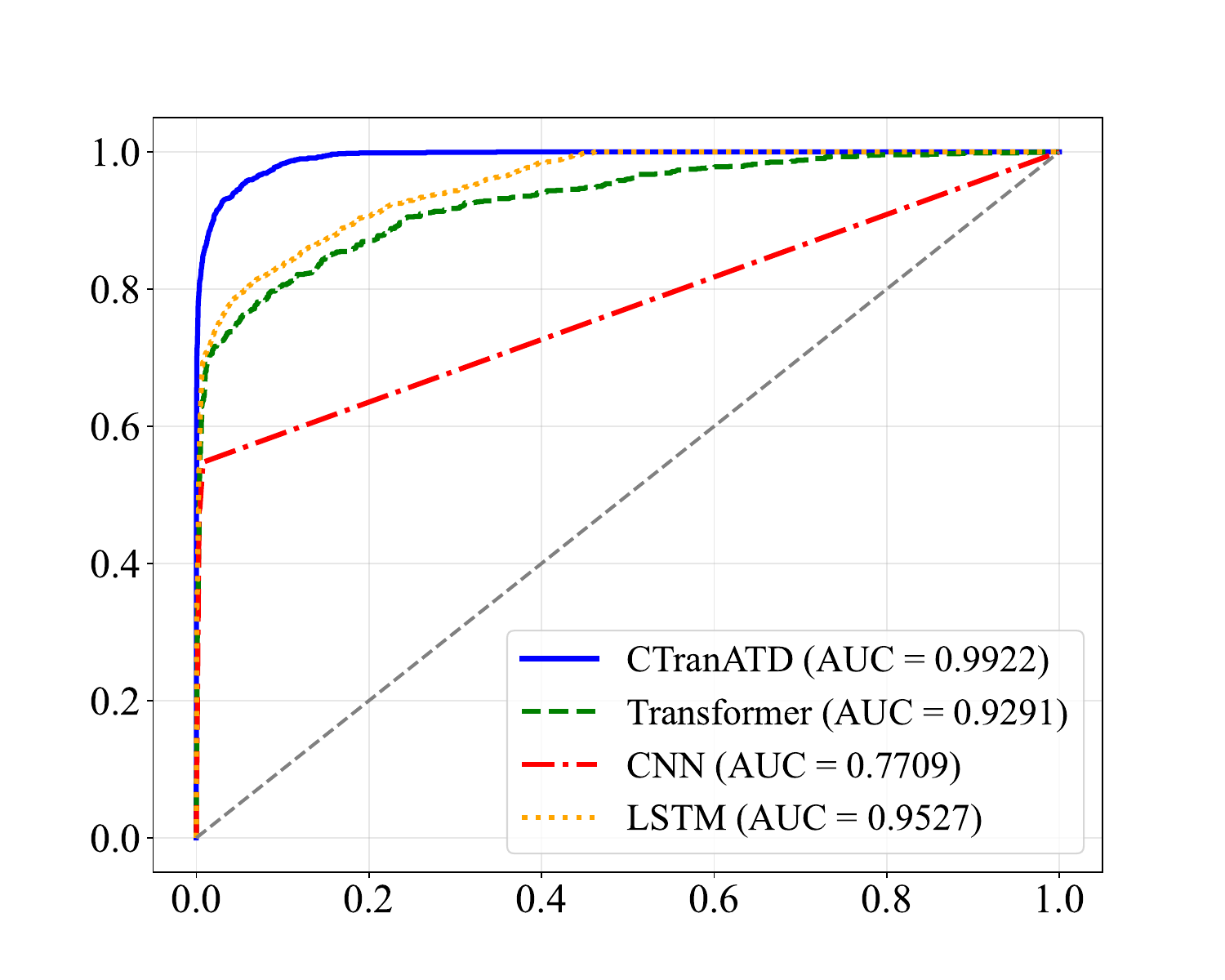}
     \label{fig:4b}
     }
     \vspace{-4mm}
     \quad
     \subfloat[ROC curves for DDoS.]
     {
	\includegraphics[width=0.60\linewidth]{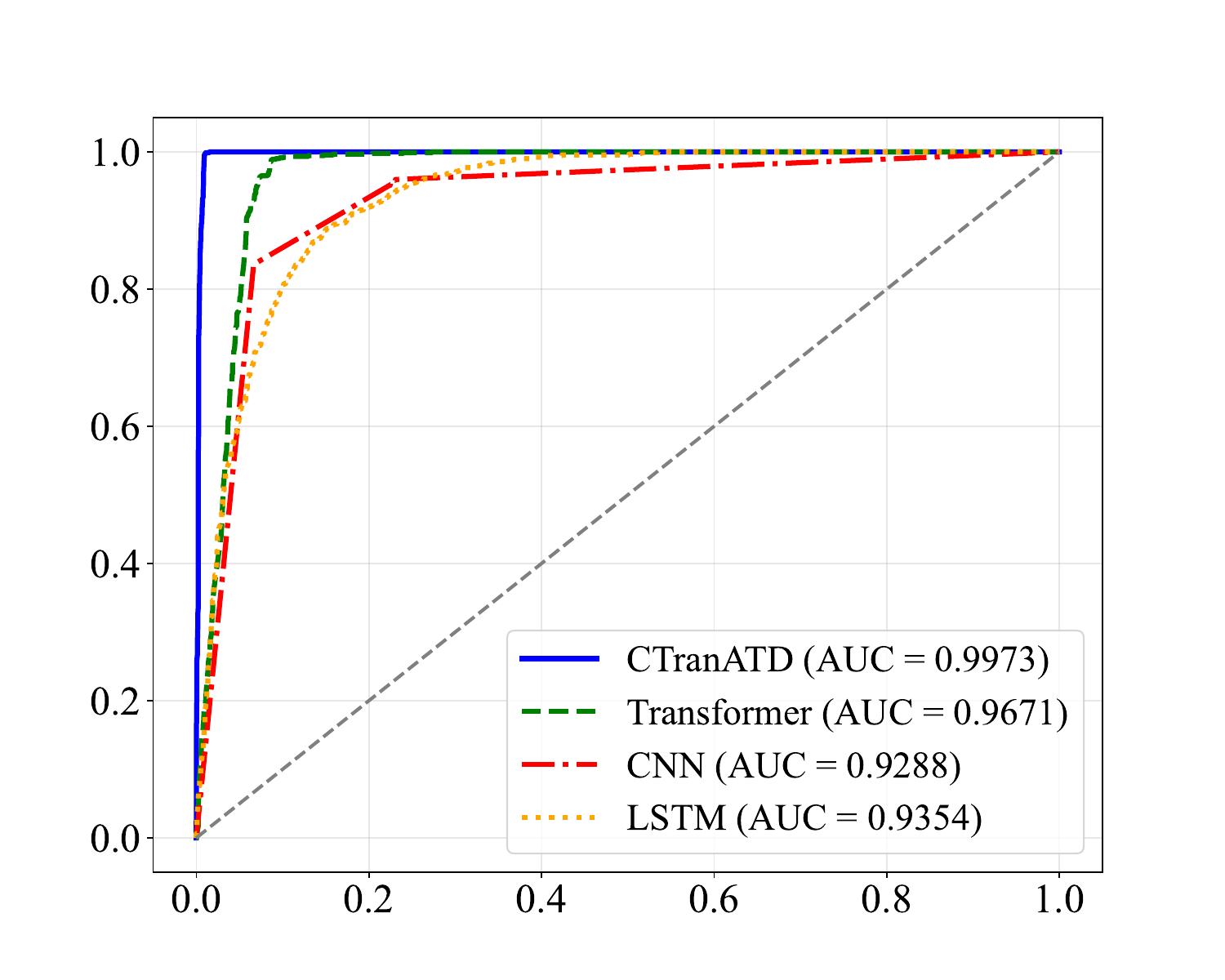}
     \label{fig:4c}
     }
     \captionsetup{font=small}
	\caption{\raggedright ROC curves of different algorithms.}
	\label{fig:ROC}
\end{figure}

Fig. \ref{fig:results} shows the performance of the anomaly detection 
model under three attacks, which are the average results 
obtained after predicting hundreds times of selections from the dataset. 
Besides, four indexes  
of the DDoS attack reach more than 0.98, and the AUC index is 
as high as 0.99, which indicates that the CTranATD can perform well 
under DDoS in anomaly detection. DoS and Portscan have 
high accuracies and AUCs more than 0.98, while the Recall and F1-score 
perform not well. It is caused 
by the imbalance of CICIDS2017 dataset. Specifically, the normal traffic accounts for 
the majority under DoS and Portsacn attacks, which leads to 
a bias in the model training. However, it has no influence on the performance of the
model in general due to the high accuracy and AUC. In short, the results show the generalization 
ability and robustness of the CTranATD. 

To verify the classification results comprehensively, 
we compare the proposed CTranATD with Transformer, CNN, and LSTM 
in Fig. \ref{fig:dif met}. It is noted that 
the CTranATD has the best performance 
in all indexes, reaching an accuracy about 0.99, while 
the single CNN algorithm has the worst classification performance. 
The performance of single Transformer algorithm is slightly inferior 
to the CTranATD. Meanwhile, LSTM is better performed 
than Transformer under Portscan and DoS. 

In Fig. \ref{fig:ROC}, we compare 
the ROC curves of four algorithms under three attacks. 
From the distribution of the ROC curves, it can be seen that the CTranATD are extremely 
smooth and close to the upper left corner under each attack, which 
is consistent with the performance of AUC value, indicating the   
outstanding capability of the CTranATD in anomaly traffic detection. The performance of Transformer  
is inferior to the CTranATD under all attacks. 
The performance of LSTM is similar to Transformer, while 
CNN has the worst performance. The individual CNN algorithm has the shortcomings of 
instability, and the classification effect is also realtively bad from the ROC curve.
\vspace{2mm}

\section{Conclusions}\label{section5}
\vspace{1mm}
In this paper, we study the anomaly detection for UAV networks. 
We construct an identity authentication security architecture by combining 
the SDN and blockchain. Then, on the SDN control plane, the anomaly 
detection algorithm is designed based on the CTranATD. 
The proposed architecture and model form a complete network security system 
for environmental monitoring and data transmission tasks 
performed by UAVs for emergency rescues. The results 
are compared with CNN, Transformer, and LSTM to 
verify that the CTranATD has a good 
capability in indentifying and detecting traffic under attacks.
\vspace{1mm}
\bibliographystyle{IEEEtran}
\bibliography{REFERENCES.bib}
\end{document}